\documentclass[runningheads]{llncs}
\usepackage[utf8]{inputenc}
\usepackage{hyperref}
\usepackage{multirow}
\usepackage[table,xcdraw]{xcolor}
\usepackage{pdflscape}
\usepackage{rotating}
\usepackage{comment}
\usepackage{svg}

\usepackage{amssymb}
\usepackage{pifont}
\newcommand{\cmark}{\ding{51}}%

\usepackage{xcolor}

\title{Finding Reusable Machine Learning Components to Build Programming Language Processing Pipelines}
\date{June 2022}

\author{Patrick Flynn \inst{2,1} 
\and Tristan Vanderbruggen\inst{1} 
\and Chunhua Liao\inst{1} 
\and Pei-Hung Lin\inst{1} 
\and Murali Emani\inst{3} 
\and Xipeng Shen \inst{4}} 
\titlerunning{Reusable PLP Components}
\authorrunning{Flynn et al.}

\institute{Lawrence Livermore National Laboratory, Livermore, CA 94550, USA \and
University of North Carolina at Charlotte, NC 28223, USA \and
Argonne National Laboratory, Lemont, IL 60439, USA \and
North Carolina State University, Raleigh, NC 27695, USA}

\begin{document}

\maketitle


\begin{abstract}
Programming Language Processing (PLP) using machine learning has made vast improvements in the past few years. Increasingly more people are interested in exploring this promising field. However, it is challenging for new researchers and developers to find the right components to construct their own machine learning pipelines, given the diverse PLP tasks to be solved, the large number of datasets and models being released, and the set of complex compilers or tools involved. To improve the findability, accessibility, interoperability and reusability (FAIRness) of machine learning components, we collect and analyze a set of representative papers in the domain of machine learning-based PLP. We then identify and characterize key concepts including PLP tasks, model architectures and supportive tools. Finally, we show some example use cases of leveraging the reusable components to construct machine learning pipelines to solve a set of PLP tasks.
\let\thefootnote\relax\footnote{\scriptsize This work was performed in part under the auspices of the U.S. Department of Energy by Lawrence Livermore National Laboratory under Contract DE-AC52-07NA27344. It is based upon work supported by the U.S. Department of Energy, Office of Science, Advanced Scientific Computing Program (ASCR SC-21) under Award Number DE-SC0021293. This work used resources of the Argonne Leadership Computing Facility (ALCF), which is a DOE Office of Science User Facility supported under Contract DE-AC02-06CH1135. LLNL-CONF-837414}
\end{abstract}



\section{Introduction}
In the past decade, machine learning (ML) has made tremendous progress in solving natural language processing (NLP) tasks. This was due to a variety of factors, including the advent of Transformer models, the availability of high quality datasets, and the development of more powerful computing systems. In particular, large-scale pretrained models, such as BERT~\cite{devlin2018bert,koroteev2021bert} and GPT-3~\cite{brown2020language}, have been a strong driver of innovation in the domain of NLP. 

Similarly, programming language processing (PLP) tasks are benefiting from the availability of pretrained models and high quality datasets. The past few years saw a vast improvement in the ability of ML to perform a large number of tasks, such as code generation\cite{https://doi.org/10.48550/arxiv.2203.07814}, clone detection\cite{https://doi.org/10.48550/arxiv.2203.07814}, source-to-source translation\cite{https://doi.org/10.48550/arxiv.2006.03511}, defect correction\cite{Bo_i__2021}, code documentation\cite{https://doi.org/10.48550/arxiv.1909.09436} and so on. There is an  increasing interest in the science community to either directly reuse or expand the ML models for PLP, in order to address software engineering challenges. 

However, the rapid development of ML for PLP also brings challenges for researchers and developers who are interested in this promising field. 
First of all, a large number of different ML models and datasets are published each year. It is difficult for people, especially newcomers, to identify representative ones to get started. 
Secondly, different model architectures are used to solve different types of PLP tasks, ranging from program understanding to code generation.  It is a daunting job for people to pick the right architectures for a given task. 
Third, machine learning workflows are typically split into a set of independent, reusable, modular components that can be pipelined together to create ML models. Mastering these components requires a significant amount of time and effort. 
Finally, in many cases, the trend is to add compiler analysis into the inputs of traditional NLP models in order to improve the quality of ML models for PLP. The use of compiler tools make the entire ML pipelines more complex, resulting in more constraints to its applicability to a given task.  


To improve the findability, accessibility, interoperability and reusability (FAIRness)
of machine learning components, we search through the literature to find representative papers in the domain of machine learning-based programming language processing. 
The goal of this paper is to facilitate the reuse of components of ML pipelines so researchers or developers can easily create customized ML pipelines to solve a given task.  




We highlight the contributions of this paper as follows.
\begin{itemize}    
    \item We identify and characterize key components of the ML pipeline for PLP, including representative tasks, popular ML models, datasets, and tools.
    \item We extend the taxonomy for PLP tasks, adding code-to-analysis and semantic-matching categories of tasks.
    \item We propose a taxonomy of tokenization tools used in PLP based on compiler engineering terminology.
    \item We demonstrate how the identified components can be composed to form different pipelines to solve given tasks.
\end{itemize}

\section{Selected Publications} 
Our search for relevant papers used the following method. 
We first selected a few recent high-impact publications (ie, BERT~\cite{devlin2018bert,koroteev2021bert}) as our initial seed publications. These were either well-known in the research community or discovered by searching for specific keywords (mostly cross-product of ``ML", ``AI" with ``source code", ``code analysis", ``code understanding") using Google Scholar.
From these seed publications, we followed both older referenced publications 
 and more recent publications citing them, especially looking for (1) the model architectures that inspired the publication, (2) their experiments (downstream tasks and datasets) and (3) which models they compared against. 
To narrow down the scope further, only representative publications with compelling artifacts, including released models, supportive tools and sufficient documentation, are selected. We still evaluated a few publications associated with some models that are only available through an API, such as AlphaCode~\cite{li2022competition} and OpenAI's CodeX~\cite{chen2021evaluating}. 
We stopped the search, when no further related papers were to be collected, to build a cohesive picture of the PLP domain presented in this paper.

Out of the surveyed publications, CodeXGLUE \cite{journal:CodeXGLUE} is the most significant. It is a benchmark dataset for code understanding and generation. It aggregates both datasets and baseline models in the PLP community. These models are based on CodeBERT~\cite{feng2020codebert} and CodeGPT \cite{journal:CodeXGLUE}, variations of BERT~\cite{devlin2018bert} and GPT~\cite{radford2018improving} trained on code.

We selected SynCoBERT\cite{https://doi.org/10.48550/arxiv.2108.04556} and ProGraML\cite{cummins2021a} to be our reference models as together they cover most of the features to form our taxonomies.
SynCoBERT is an extension to CodeBERT and CuBERT \cite{https://doi.org/10.48550/arxiv.2001.00059}. It considers source code, natural language text, and the corresponding AST to enable multi-modal pretraining for code representation.
It also presents an evaluation with common code-to-code, code-to-text, and text-to-code tasks.
ProGraML 
applies graph neural networks to graph representations of LLVM IR. Most importantly, ProGraML is used to solve five traditional compiler analysis tasks, including control- and data-flow, function boundaries, instruction types, and the type and order of operands over complex programs.

\section{Taxonomies}
\label{sec:taxonomies}

A set of standard taxonomies are indispensable to enable findable, accessible, interoperable and reusable components in the domain of Programming Language Processing (PLP) using machine learning (ML). We extend CodeXGLUE's classification of PLP tasks, organize relevant model architectures, and propose a taxonomy of tokenization tools/methods.

\subsection{PLP Tasks}
\label{sec:taxonomies:tasks}

The surveyed publications used a variety of downstream PLP tasks to evaluate the capability of their models. Most of these tasks fall clearly under one of the four categories introduced by CodeXGLUE: Code-to-Code, Code-to-Text, Text-to-Code, and Text-to-Text. Clearly, these categories are defined in terms of the input and output modalities. 

However, the authors of CodeXGLUE only considered \textit{code} and \textit{text}, which are sequences of programming language and natural language tokens.  It means that tasks where the output is a scalar value (classification and regression tasks) cannot be easily categorized. For example, defect detection aims at classifying code samples between correct and defective, it is a classification task. Clone detection and code search aim at predicting semantic matching between two sequences. ProGraML and its predecessor NCC use tasks that target scalar values: OpenCL device mapping, OpenCL thread coarsening factor, and algorithm classification.

\begin{figure}[htbp]
  \vspace{-\baselineskip}
  \centering
  \includegraphics[width=1.0\columnwidth]{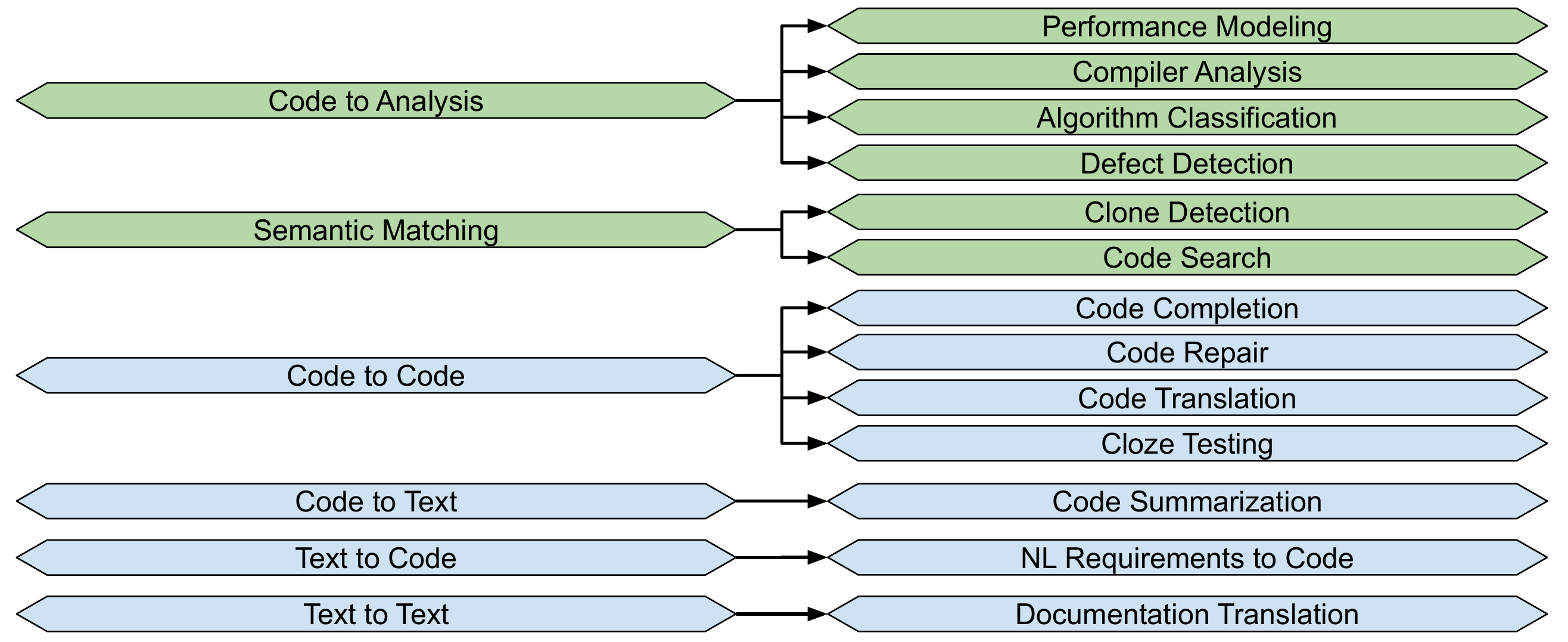}
  \caption{Our taxonomy of tasks based on CodeXGLUE's categorization}
  \label{fig:taxo:tasks}
\end{figure}

Given the limitation of CodeXGLUE's task categorization, we propose to add two new categories of PLP tasks: code-to-analysis and semantic-matching. This is illustrated in Figure~\ref{fig:taxo:tasks}.

The \textit{Code-to-Analysis}  category groups tasks taking code as input and generating results that are neither text nor code. 
We create four subcategories in this category:
1) Performance modeling either predicts a runtime metric of the code or a configuration that optimizes the said metric.
2) Compiler analysis includes traditional static compiler analysis tasks, particularly the reachability, dominator trees, and liveness analysis from ProGraML. 
3) Algorithm detection is similar to clone detection, but we narrowly define it as detecting the use of a finite set of algorithms in a code.
4) Defect detection may use previous code commits with information about resolved vulnerabilities to predict detects in new code.

Semantic-matching includes both clone detection \cite{6976121} and code search \cite{https://doi.org/10.48550/arxiv.1909.09436} tasks. The current state-of-the-art approaches use deep-learning to formulate search problems as the evaluation of the semantic similarity between a ``query'' and each element of a set of ``candidates''. The result of this formulation are models that take a pair of code-code (used by clone detection) or text-code (used by code search) sequences as inputs and predict a similarity score for the pair. 

The Code-to-Code, Code-to-Text, Text-to-Code, and Text-to-Text categories are the same as described in CodeXGlue. The \textit{Code-to-Code} category includes code completion, code repair, code translation, and cloze testing (occlusion test or ``filling the blank'' colloquially). Code completion aims to predict the next code token or the next section of code. Code repairs aims at reproducing a code without an existing defect. These defects are found by mining git repositories for very simple commits with descriptive messages of the issue being fixed. Code translation aims to convert a body of text from one programming language to another. Code-to-text includes code summarization, which aims to produce documentation for a given piece of code. Text-to-code includes generating code from natural language requirements. Text-to-text includes document translation across natural languages.

Finally, we must mention self-supervised tasks that are used to pretrain models. We describe a few of these in more detail when illustrating BERT and SynCoBert. We particularly look into Masked Language Modeling (MLM) which is a form of Cloze Testing, Identifier Prediction (IP), and Edge Prediction (EP). The interesting fact about these tasks is how much they are tied to the tokenization tools that we introduce in Section~\ref{sec:taxonomies:tools}.

In Table~\ref{tbl:tasks_datasets}, PLP tasks are associated with pretrained models that are used to support downstream tasks and the datasets from which they can be derived. This table can also be used to guide users to easily find the right models and datasets for a given task.  


\begin{table}[htb]
\resizebox{12.5cm}{!}{
\begin{tabular}{|l
            |l|l|l|l|l|l|l|l|l|l|l|l
            |l|l|l|l|l|l|l|l|l|l|l|l|l|l|l|l|l|l|l|l|l|l|}
    \hline
    \multicolumn{1}{|c}{Tasks} & \multicolumn{12}{|c|}{Models} & \multicolumn{22}{c|}{Datasets} \\
    \hline
    & \parbox[t]{2mm}{\rotatebox[origin=c]{90}{PrograML\cite{cummins2021a}}}
    & \parbox[t]{2mm}{\rotatebox[origin=c]{90}{SynCoBERT\cite{https://doi.org/10.48550/arxiv.2108.04556}}}
    & \parbox[t]{2mm}{\rotatebox[origin=c]{90}{PLABART\cite{ahmad-etal-2021-unified}}}
    & \parbox[t]{2mm}{\rotatebox[origin=c]{90}{CodeT5\cite{wang-etal-2021-codet5}}}
    & \parbox[t]{2mm}{\rotatebox[origin=c]{90}{Code-MVP\cite{https://doi.org/10.48550/arxiv.2205.02029}}}
    & \parbox[t]{2mm}{\rotatebox[origin=c]{90}{TreeBERT\cite{pmlr-v161-jiang21a}}}
    & \parbox[t]{2mm}{\rotatebox[origin=c]{90}{ContraCode\cite{jain2020contrastive}}}
    & \parbox[t]{2mm}{\rotatebox[origin=c]{90}{GraphCodeBERT\cite{https://doi.org/10.48550/arxiv.2009.08366}}}
    & \parbox[t]{2mm}{\rotatebox[origin=c]{90}{CoTexT\cite{phan2021cotext}}}
    & \parbox[t]{2mm}{\rotatebox[origin=c]{90}{CodeBERT\cite{feng2020codebert}}}
    & \parbox[t]{2mm}{\rotatebox[origin=c]{90}{CodeGPT\cite{journal:CodeXGLUE}}}
    & \parbox[t]{2mm}{\rotatebox[origin=c]{90}{MICSAS\cite{journal:CodeXGLUE}}}
    
    & \parbox[t]{2mm}{\rotatebox[origin=c]{90}{CodeSearchNet\cite{https://doi.org/10.48550/arxiv.1909.09436}}} 
    & \parbox[t]{2mm}{\rotatebox[origin=c]{90}{DeepTune OpenCL\cite{cummins2017b}}}
    & \parbox[t]{2mm}{\rotatebox[origin=c]{90}{POJ-104\cite{mou2016convolutional}}}
    & \parbox[t]{2mm}{\rotatebox[origin=c]{90}{BigCloneBench\cite{6976121}}}
    & \parbox[t]{2mm}{\rotatebox[origin=c]{90}{Defects4J\cite{10.1145/2610384.2628055}}}
   & \parbox[t]{2mm}{\rotatebox[origin=c]{90}{Devign\cite{https://doi.org/10.48550/arxiv.1909.03496}}}
   & \parbox[t]{2mm}{\rotatebox[origin=c]{90}{DeepDataFlow\cite{cummins2021a}}}
   & \parbox[t]{2mm}{\rotatebox[origin=c]{90}{DeepTyper\cite{10.1145/3236024.3236051}}}
   & \parbox[t]{2mm}{\rotatebox[origin=c]{90}{PY150\cite{10.1145/3022671.2984041}}}
   & \parbox[t]{2mm}{\rotatebox[origin=c]{90}{Github Java Corpus\cite{githubCorpus2013}}} 
   & \parbox[t]{2mm}{\rotatebox[origin=c]{90}{Tufano's dataset\cite{10.1145/3340544}}}
   & \parbox[t]{2mm}{\rotatebox[origin=c]{90}{Nguyen's dataset\cite{7372046}}}
   & \parbox[t]{2mm}{\rotatebox[origin=c]{90}{CodeTrans\cite{10.5555/3327144.3327180}}}
   & \parbox[t]{2mm}{\rotatebox[origin=c]{90}{AdvTest\cite{https://doi.org/10.48550/arxiv.1909.09436}}}
   & \parbox[t]{2mm}{\rotatebox[origin=c]{90}{CONCODE\cite{https://doi.org/10.48550/arxiv.1808.09588}}}
   & \parbox[t]{2mm}{\rotatebox[origin=c]{90}{CosQA\cite{https://doi.org/10.48550/arxiv.2105.13239}}}
   & \parbox[t]{2mm}{\rotatebox[origin=c]{90}{CoNaLa\cite{yin2018mining}}}
   & \parbox[t]{2mm}{\rotatebox[origin=c]{90}{DeepCom\cite{hu_li_xia_lo_jin_2019}}}
   & \parbox[t]{2mm}{\rotatebox[origin=c]{90}{Python8000\cite{https://doi.org/10.48550/arxiv.2105.12655}}} 
   & \parbox[t]{2mm}{\rotatebox[origin=c]{90}{Google Code Jam \cite{GoogleCodeJam}}} 
   & \parbox[t]{2mm}{\rotatebox[origin=c]{90}{CodeContests\cite{li2022competition}}} 
    & \parbox[t]{2mm}{\rotatebox[origin=c]{90}{CodeXGlue\cite{journal:CodeXGLUE}}}
    \\ \hline
Performance Modeling & \cmark &  &  &  &  &  &  &  &  &   &  &
&  & \cmark &  &  &  &  &   &  &  &  &  &  &  &  &  &  &  &  &  & & &
\\ \hline

Algorithm Classification & \cmark &  &  &  &  &  &  &  &  &  &  &
&  &  & \cmark &  &  & \cmark &  &  &  &  &  &  &  &  &  &  &  &  &  & & &
\\ \hline

Defect Detection         &  & \cmark & \cmark & \cmark & \cmark &  &  &  & \cmark &  &  &
& &  & & \cmark & \cmark & \cmark & &  &  &  &  &  &  &  &  &  &  &  &  & & & \cmark
\\ \hline

Compiler Analyses        & \cmark &  &  &  &  &  &  &  &  &  &  &
&  &  &  &  &  &  & \cmark & \cmark &  &  &  &  &  &  &  &  &  &  &  & & & 
\\ \hline

Code Completion          &  &  &  &  &  &  & \cmark &  &  &  & \cmark &
&  &  &  &  &  &  & &  & \cmark & \cmark &  &  &  &  &  &  &  &  &  & & & \cmark
\\ \hline

Code Repair              &  &  &  & \cmark &  &  &  & \cmark & \cmark &  &  &
&  &  &  &  &  &  &  &  &  &  & \cmark &  &  &  &  &  &  &  &  & & & \cmark

\\ \hline
Code Translation         &  & \cmark & \cmark & \cmark &  &  &  & \cmark &  &  &  &
&  &  &  &  &  &  &  &  &  &  &  & \cmark & \cmark &  &  &  & \cmark &  &  &  \cmark & \cmark & \cmark

\\ \hline
Cloze Testing            &  &  &  &  &  &  &  &  &  & \cmark &  &
& \cmark &  &  &  &  &  &  &  &  &  &  &  &  &  &  &  &  &  &  & & & \cmark
\\ \hline

Text-to-Code Generation  &  &  & \cmark & \cmark &  &  &  &  & \cmark &  & \cmark &
& \cmark &  &  &  &  &  &  &  &  &  &  &  &  &  & \cmark & \cmark &  &  &  & & & \cmark
\\ \hline

Code Summarization       &  &  & \cmark & \cmark &  & \cmark & \cmark &  & \cmark & \cmark &  &
& \cmark &  &  &  &  &  &  &  &  &  &  &  &  &  &  &  &  & \cmark &  & & & \cmark
\\ \hline

Document Translation     &  &  &  &  &  &  &  &  &  &  &  & 
&  &  &  &  &  &  & &  &  &  &  &  &  &  &  &  &  &  &  & & & \cmark
\\ \hline

Code Search              &  & \cmark &  &  & \cmark  &  &  & \cmark &  & \cmark &  &
& \cmark &  &  &  &  &  &  &  &  &  &  &  &  & \cmark &  &  &  &  &  & & & \cmark
\\ \hline

Clone Detection &  & \cmark & \cmark & \cmark & \cmark &  & \cmark & \cmark &  &  &  & \cmark
&  &  & \cmark & \cmark &  &  &  &  &  &  &  &  &  &  &  &  &  &  & \cmark & \cmark &  & \cmark
\\ \hline

\end{tabular}
} 
\caption{Association between PLP tasks and publications (focused on either model or dataset. Only explicit associations are marked in this table since some models and datasets may be adapted to serve other tasks. )}
\vspace{-2.0\baselineskip}
\label{tbl:tasks_datasets}
\end{table}

\subsection{Model Architectures}
\label{sec:taxonomies:models}

We have seen that models can be trained to solve a number of different tasks. In this section, we categorize the various neural architectures used to solve PLP tasks. Being aware of the various architectures and how they can be composed is essential to leveraging the pretrained models. Indeed, often the input-output of the pretraining does not match the task that we wish to specialize it for. In this case, parts of the pretrained model are discarded and new components are trained from scratch. This leads to some confusion, as a given \textit{model} is often used to refer to either its neural architecture or a trained instance of the said architecture.

\begin{figure}[htbp]
  \vspace{-\baselineskip}
  \centering
  \includegraphics[width=1.0\columnwidth]{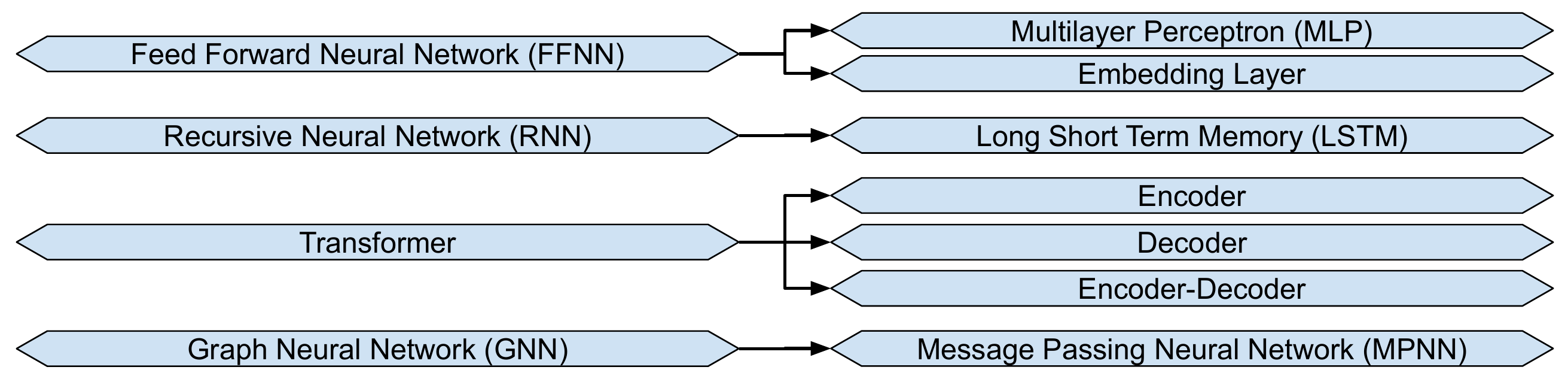}
  \caption{Taxonomy of architectures based on the modality they operate on.}
  \label{fig:taxo:models}
  \vspace{-\baselineskip}
\end{figure}

We present a coarse taxonomy of the different architecture we have encountered, as shown in Figure~\ref{fig:taxo:models}. Our taxonomy aims to be accessible to non-expert in machine learning. In comparison, fine-grained taxonomies could be confusing to users with too much low-level details. They may not be able to accommodate fast evolving machine learning technologies. 
It is based on the modality on which the architecture operates. This categorization is common for a general purpose presentation of Neural Networks. We redirect the reader toward \cite{https://doi.org/10.48550/arxiv.2108.05542,sarker2021deep} and \cite{https://doi.org/10.48550/arxiv.2005.03675} for taxonomies of NLP models and graph models, respectively.

\begin{figure}[htbp]
  \vspace{-.5\baselineskip}
  \centering
  \includegraphics[width=\columnwidth]{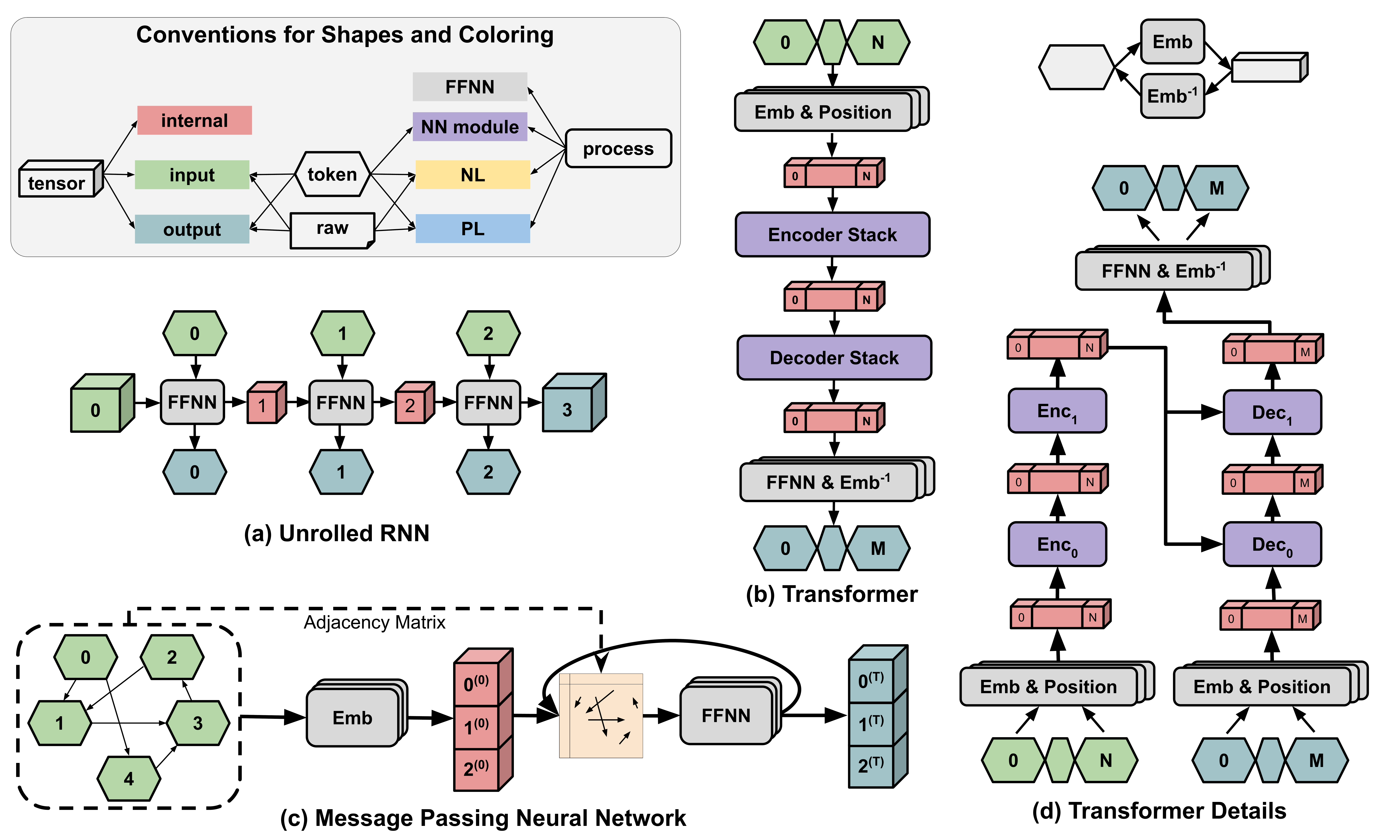}
  \caption{
    Architectural building blocks and drawing conventions.
  }
  \label{fig:models}
  \vspace{-\baselineskip}
\end{figure}

Figure~\ref{fig:models} illustrates some common building blocks of various architectures. The top left diagram shows conventions for shapes and coloring.
The figure also includes the most basic RNN (a), Message Passing Neural Network (c) and the coarse-grained details of the Transformer (d) to illustrate the fundamental difference between these architectures. 
While RNN must propagate information from the first token to the last, Transformers see all the input tokens and the decoder sees all the previous output tokens (and the encoded inputs).
A full Transformer (b) is presented to illustrate BERT and SynCoBERT in Section~\ref{sec:pipelines:examples}. The Transformer is modeled with four blocks: token and positional embeddings, encoder-stack, decoder-stack, and finally embedding reversal. 

Feed Forward Neural Networks (FFNN) operate on tensors. Their neurons' connections form a DAG. Common FFNNs are the multilayer perceptron (MLP) or a simple embedding layer. FFNNs are the building blocks of deep-learning.
Recursive Neural Networks (RNN) operate on sequences (of tensors). RNNs have an initial state (tensor) that is updated for each element of the input sequence. After each update, an output can be produced while the final state can be used for sequence understanding. Long Short Term Memory (LSTM) is an advanced RNN architecture.

Transformers have changed the landscape of deep-learning quite radically. Initially the attention mechanism, which is the base of the transformer architecture, was used as part of RNN architectures. However, since its introduction in the aptly named ``Attention is All You Need'' paper in 2017 \cite{https://doi.org/10.48550/arxiv.1706.03762}, Transformers have replaced RNNs for language modeling tasks.
Transformers have also shown the ability to outperform convolutional neural networks (CNN) for image processing \cite{https://doi.org/10.48550/arxiv.2010.11929,https://doi.org/10.48550/arxiv.2005.12872} and text-to-image \cite{https://doi.org/10.48550/arxiv.2206.10789}.

The Transformer architecture uses attention,
a deep-learning mechanism, whereas the dot-product of keys and queries measures the \textit{attention} that should be given to a value. The nature of the attention mechanism makes transformers a set-to-set architecture. However, by simply adding a positional embedding to each token's embedding, Transformers act as sequence-to-sequence architectures.

\begin{figure}[htbp]
  \centering
  \includegraphics[width=\columnwidth]{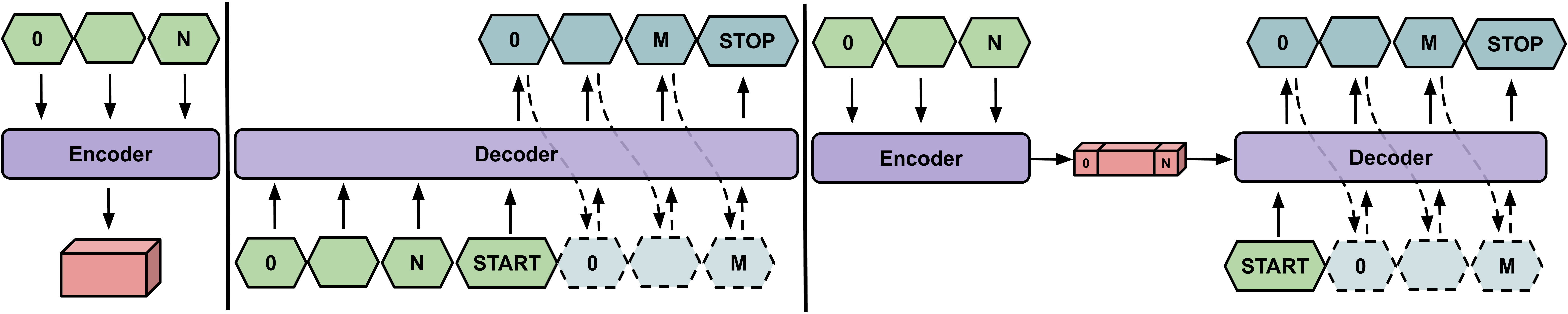}
  \caption{Left-to-right: a) an encoder produces a tensor from a sequence of tokens, b) a decoder-based model sees an input sequence and autoregressively generates the output sequence, and c) an encoder-decoder uses the tensor from the encoder to parameterize an autoregressive decoder. (See color legend in figure~\ref{fig:models})}
  \label{fig:encoder-decoder}
  \vspace{-.5\baselineskip}
\end{figure}

In our taxonomy, we highlighted three sorts of transformers: Encoder, Decoder and Encoder-Decoder. In the original Transformers~\cite{vaswani2017attention}, both an encoder stack and a decoder stack are used to produce the output as depicted in Figure~\ref{fig:models}d. However, the architecture can be split as shown in Figure~\ref{fig:models}b. Following this realization, both encoder-only and decoder-only transformers have been devised. Bidirectional Encoder Representations from Transformers (BERT)~\cite{devlin2018bert}, and Generative Pretrained Transformer (GPT)~\cite{radford2018improving} are respective examples of the Encoder and the Decoder architecture. This separation between encoder and decoder components is common in ML. We illustrate in Figure~\ref{fig:encoder-decoder} how this paradigm applies to the processing of sequence of tokens.

Graph Neural Networks (GNN) generalize convolutional neural networks (CNN) from the grid of an image to the unstructured mesh defined by a graph. We focused on Message Passing Neural Networks (MPNN), a recent implementation of the paradigm which offers a lot of flexibility while having good computational performance (by exploiting the sparsity of the adjacency matrix).


Finally, Mixture-of-Expert (MoE) models are increasingly being used to scale large language models efficiently. The modern approach to MoE consists of sparsely activated models which minimizes resource utilization. For example \cite{deepspeed} implemented the MoE-based model to scale a 700 million-parameter dense model from 1.8 billion parameters with eight experts to 10 billion parameters using 64 experts, with no impact on the model convergence. The building blocks within the core transformer model are still the same and it has little influence on reusability.

\subsection{Tokenization Tools}
\label{sec:taxonomies:tools}

One of the significant changes in the past few years is the integration of compiler-based analysis results in PLP to enrich the representation of code. In this section, we categorize the tokenization tools used in PLP as they are different from the traditional NL Tokenization tools (such as NLTK, spaCy, TextBlob, and WordPiece).

\begin{figure}[htbp]
  \vspace{-\baselineskip}
  \centering
  \includegraphics[width=\columnwidth]{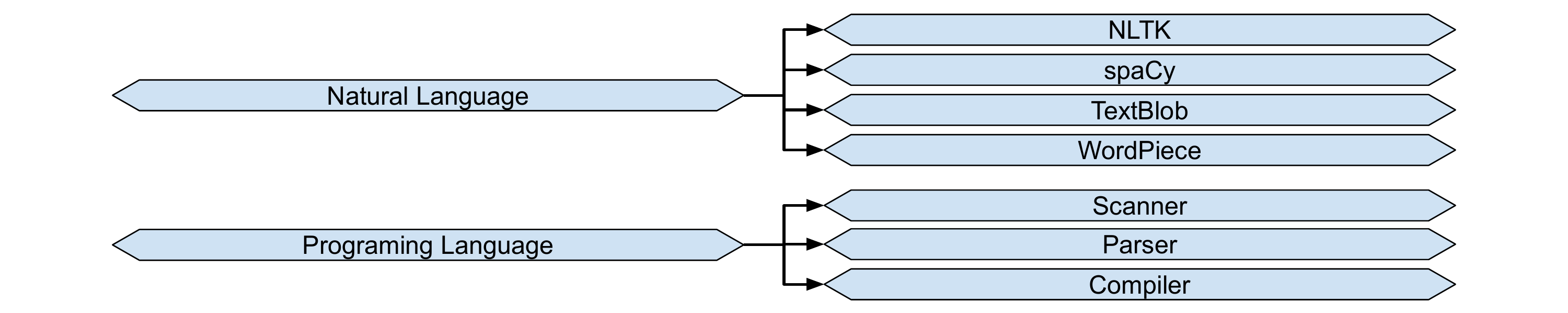}
  \caption{Our taxonomy of tools.}
  \label{fig:taxo:tools}
  \vspace{-\baselineskip}
\end{figure}

\begin{figure}[htbp]
  \centering
  \includegraphics[width=1.0\columnwidth]{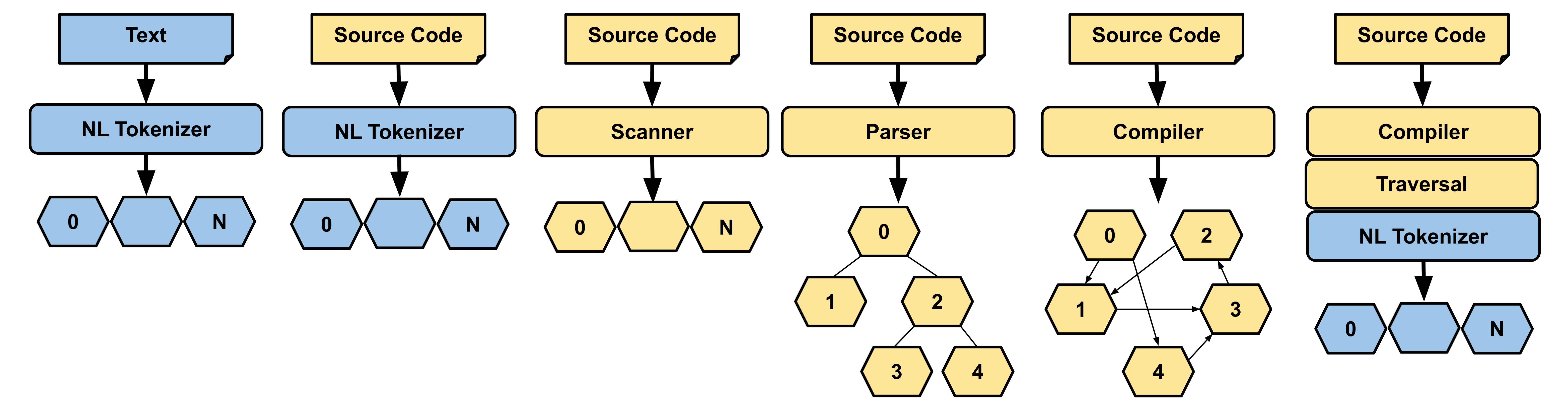}
  \caption{Tokenization building block for the pipeline (See color legend in figure~\ref{fig:models})}
  \label{fig:tools}
  \vspace{-.5\baselineskip}
\end{figure}


We organize Programming Language Tokenization tools based on three traditional stages in compiler frontends: lexical analysis, syntactic analysis, and semantic analysis (or elaboration).
The names \textit{scanner} and \textit{parser} are typically used for the first two, while the more general term \textit{compiler} is used for the last one. Although \textit{compiler} generally refers to all three of these tools together, we use it here because it includes both semantic analysis and other advanced analyses that are compiler-specific, while differentiating from the \textit{scanner} and \textit{parser}.
A \textit{scanner} turns an input into a list of tokens whereas a \textit{parser} converts the list of tokens, generated by a \textit{scanner}, into a data structure, like a tree, according to the grammatical rules.
This categorization differentiates the tools based on the output modalities: a sequence of tokens for \textit{scanners}, a tree of tokens for \textit{parsers}, and a graph of tokens for \textit{compilers}. We illustrate the different scenarios in Figure~\ref{fig:tools}. We found that in many cases the right most construct is used to leverage compiler tools. In this case, the tree or graph is traversed using a predetermined \textit{traversal}. The resulting sequence of tokens is then joined and tokenized using a NL tokenizer. We discuss in Section~\ref{sec:pipelines:examples} how self-supervised pretraining tasks are constructed using additional information from the PL tokenizers.

In Table~\ref{tbl:table_tools}, we summarize the tokenizers used by several of the publications that we surveyed.

\begin{table}[htbp]
\centering
    \begin{tabular}{|c|c|c|c|c|c|} 
     \hline
     \rowcolor[HTML]{EFEFEF} 
     \textbf{Model} & \textbf{Tools} & \textbf{NL} & \textbf{Scanner} & \textbf{Parser} & \textbf{Compiler} \\
     \hline
     NCC            & LLVM &  &  &  & \cmark \\
     \hline
     PrograML       & LLVM &  &  &  & \cmark \\
     \hline
     CodeBERT       & WordPiece & \cmark &  &  &  \\
     \hline
     GraphCodeBERT  & tree-sitter & \cmark &  & \cmark &  \\
     \hline
     SynCoBERT      & tree-sitter & \cmark &  & \cmark &  \\
     \hline
     PLABART        & Custom & \cmark & \cmark &  &  \\
     \hline
     TreeBERT       & tree-sitter & \cmark &  & \cmark &  \\
     \hline 
     ContraCode     & BabelJS (compiler) &  &  & \cmark & \cmark \\
     \hline
     CoTexT         & Custom & \cmark & \cmark &  &  \\
     \hline
     CodeT5         & tree-sitter & \cmark & \cmark & \cmark &  \\
     \hline
     Code-MVP       & tree-sitter, Scalpel, Custom & \cmark & \cmark & \cmark & \cmark \\
     \hline
     MICSAS       & Custom &  & \cmark & \cmark &  \\

     \hline
    \end{tabular}
    \caption{Models \& Tools with input types}
    \vspace{-1.0\baselineskip}
    \label{tbl:table_tools}
\end{table}

\section{PLP Pipelines}
\label{sec:pipelines}

In this section, we demonstrate how to assemble a PLP pipeline.
It is built by composing model architectures to produce the targets specified by tasks given the tokens generated by tokenization tools.
The pipeline models how raw representations flow through different components.
Given the prohibitive cost of training large scale models, pretrained models can be the only way to solve complex downstream tasks. 
We look at two representative pretraining pipelines, and then model how it can be used to solve our downstream tasks.

\subsection{Pretraining Pipelines}
\label{sec:pipelines:examples}
\label{sec:pipelines:pretraining}

\begin{figure}[htbp]
  \vspace{-\baselineskip}
  \centering
  \includegraphics[width=\columnwidth]{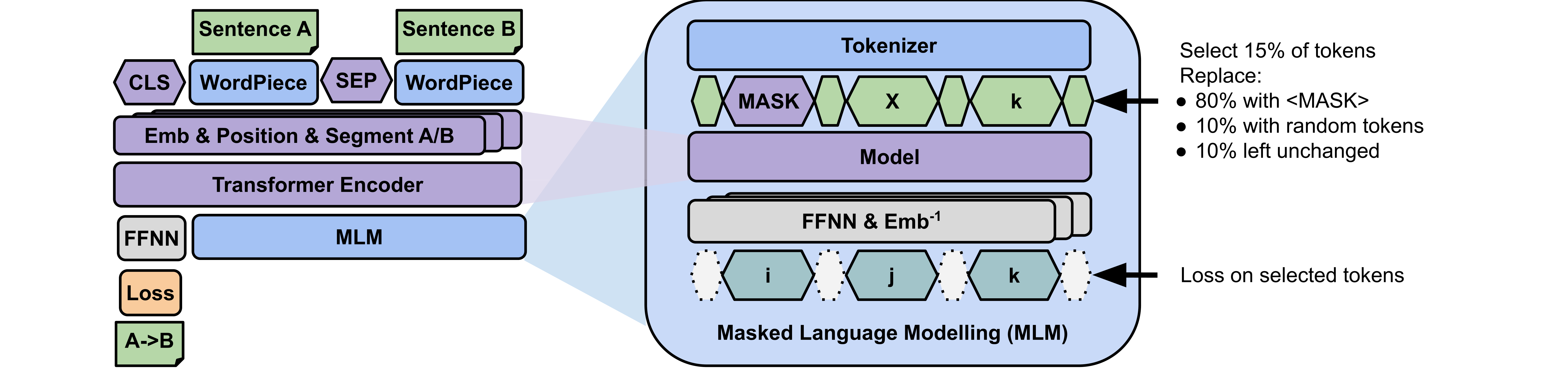}
  \caption{Self-supervised pretraining pipeline of BERT. (See color legend in figure~\ref{fig:models})}
  \label{fig:bert}
  \vspace{-\baselineskip}
\end{figure}

Figure~\ref{fig:bert} illustrates the pretraining pipeline of BERT \cite{devlin2018bert}. On the left, we show a high-level view of the pipeline.
It takes two sentences as inputs which are tokenized using WordPiece, prefixed with the classification token, and separated using a separator token.
These tokens are encoded, and positional and segment embeddings are added.
A transformer encoder stack is used to produce bidirectional embeddings of each token.
The embedding of the classification token, referred to as ``classification head'', is used to predict whether sentence A directly precedes sentence B.
All other embeddings are trained using the Masked Language Model (MLM) self-supervised tasks which we illustrate in the right part of Figure~\ref{fig:bert}.

CodeBERT \cite{feng2020codebert} uses the same pipeline with pairs of text and code instead of sentences. The classification head is made to predict whether the text and code are related. They used CodeSearchNet \cite{https://doi.org/10.48550/arxiv.1909.09436} to provide matching pairs of text and code.

\begin{figure}[htbp]
  \centering
  \includegraphics[width=\columnwidth]{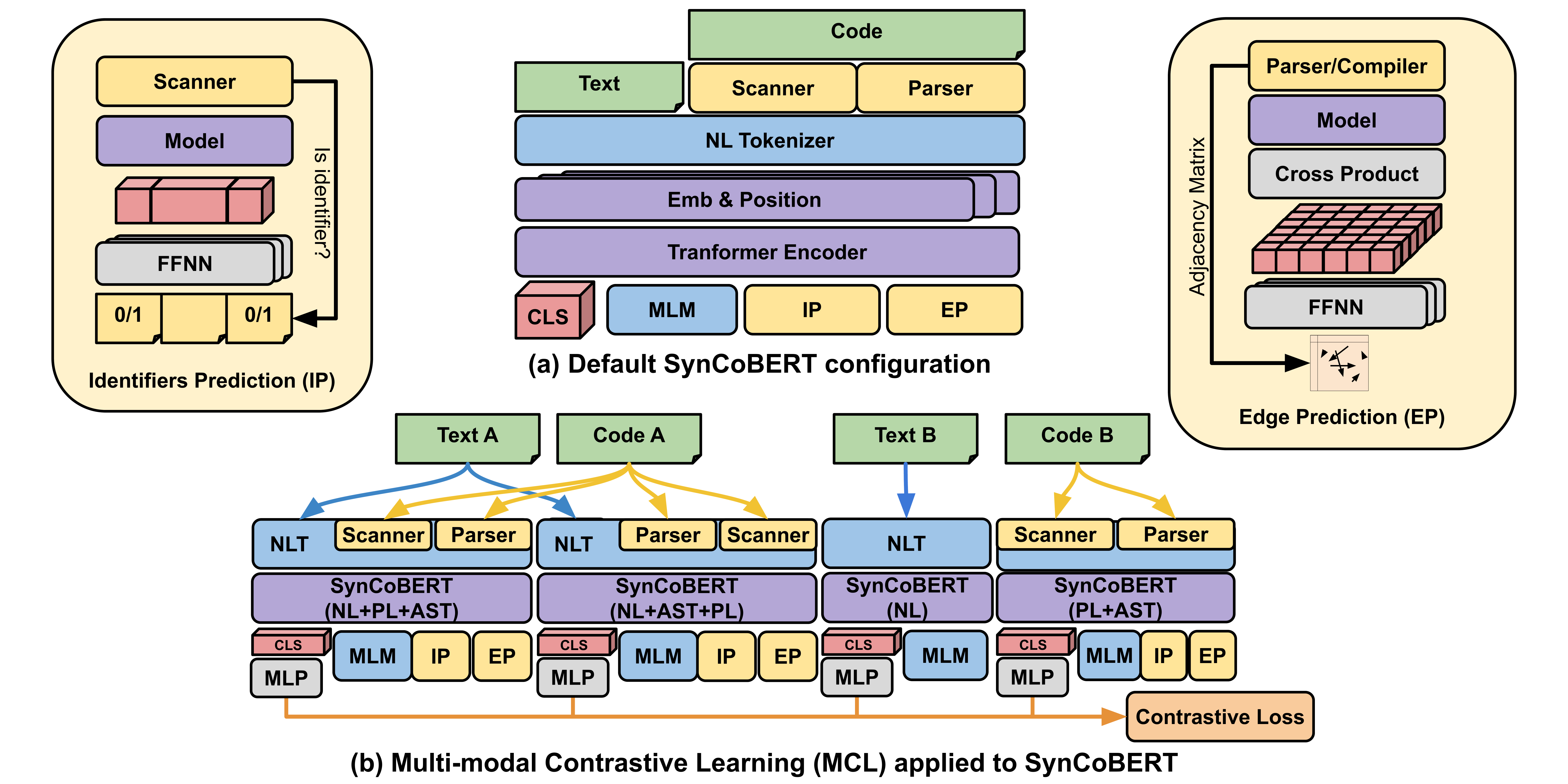}
  \caption{SynCoBERT pretraining pipeline (See color legend in figure~\ref{fig:models})
  }
  \label{fig:syncobert}
\end{figure}

SynCoBERT \cite{https://doi.org/10.48550/arxiv.2108.04556} is a BERT-like PLP model which demonstrates a complex PLP pipeline, as shown in Figure~\ref{fig:syncobert}. During pretraining, SynCoBERT inputs can be text and/or code. The code is processed into scanned tokens and AST tokens (resulting from a pre-order depth-first traversal of the tree). The final text tokens, code tokens, and AST tokens are produced using a NL tokenizer. The different sequences are separated by a special token but there is no segment embedding. SynCoBERT is pretrained using multiple self-supervised tasks: Masked Language Model (MLM), Identifier Prediction (IP), and Edge Prediction (EP). IP uses information from the scanner to determine which of the tokens are part of an identifier (NL tokenizer can split words). Then, it trains a FFNN to predict from the encoding of a token whether or not it is part of an identifier. EP uses the adjacency matrix of the AST as a target. It trains a FFNN to predict whether there is an edge between two tokens from their embeddings. Both IP and EP aims at making the encoder understand the syntax of code and AST, and add that information into the embedding. Finally, SynCoBERT uses multi-modal contrastive learning (MCL) to train a multi-layer perceptron (MLP) on the classification head. This approach helps the encoder to produce a representation that distinguishes and correlates the different inputs.

\subsection{Pipeline Specialization}
\label{sec:pipelines:specialisation}
We will now devise how we could leverage the pretrained CodeBERT or SynCoBert to solve our downstream tasks. The two example downstream tasks are: (1) code similarity analysis on a dataset of OpenMP codes, and (2) listing the fully qualified names of all declarations given a piece of C++ code.

The first task is to analyze the codes from DataRaceBench~\cite{liao2017dataracebench}, a benchmark suite of both correct and defective OpenMP codes. 
This benchmark is designed to systematically and quantitatively evaluate the effectiveness of data race detection tools. We want to use ML to identify redundant source code and identify potential gaps in coverage of typical code patterns. This is a specialized code similarity analysis task.

\begin{figure}[htbp]
  \centering
  \includegraphics[width=\columnwidth]{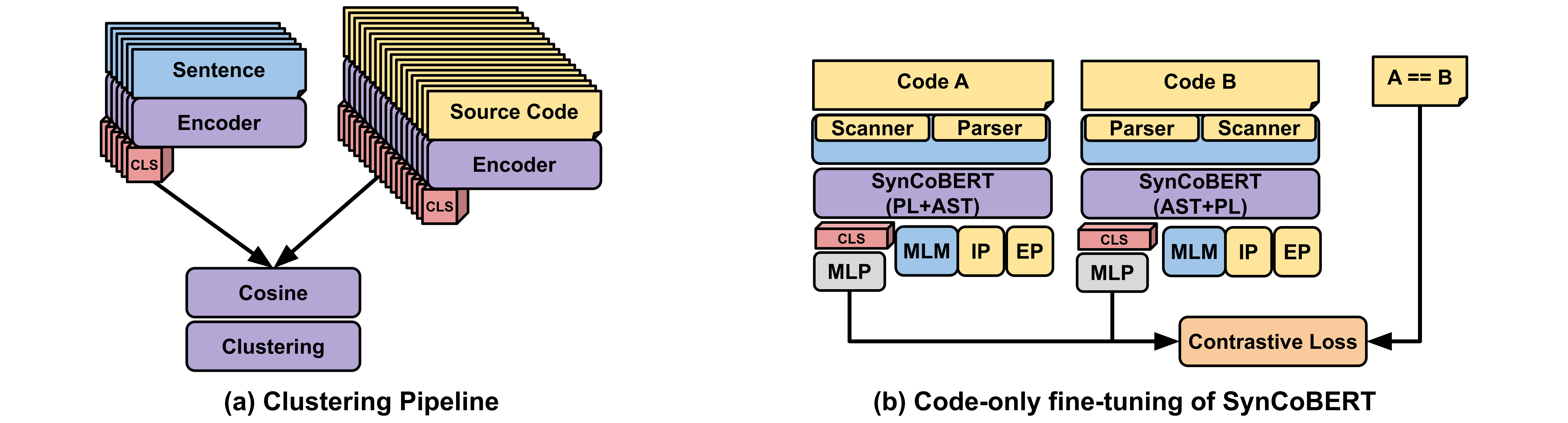}
  \caption{
     This figure illustrates the zero-shot approach to information discovery in a dataset of codes.
     To the left, we illustrate how a pretrained model can be used to produce embedding of codes and relevant sentences.
     These embeddings are then compared with pairwise cosine and clustered using conventional techniques.
     To the right, we illustrate how we could fine-tune SynCoBERT in the absence of text-code pairs. (See color legend in figure~\ref{fig:models})
  }
  \label{fig:databenchrace}
  \vspace{-.5\baselineskip}
\end{figure}

For this task, we will use zero-shot or few-shot learning pipeline which has become practical with the advent of large scale pretrained models. In Figure~\ref{fig:databenchrace}a, the pipeline applies pretrained BERT-like models to each of the samples and collects the embedding on the classification head. It then uses the same models to embed relevant text (we consider mining the publication for relevant sentences). After that, the pipeline applies clustering techniques based on cosine distance to analyze the dataset. The expectation here is to find sentences clustered alongside the code providing descriptions of the clusters.
In the few-shot variation, we would first fine-tune the encoder on this dataset. This will not be as applicable for CodeBERT pretraining pipeline as the classification head requires paired inputs of text and code to be trained. For SynCoBERT, we could fine tune with paired code-code by only using (PL-AST) vs (AST-PL) in the MCL task. We are not sure how this code-only fine tuning would affect the quality of the text embeddings used for the reference sentences.

The second task is to list all declarations in C++ code (fully-qualified). This is the simplest task we could devise for a lexical analyzer (parser). For this task, we are using automatically generated C++ source codes. Targets are produced by the code generator.


\begin{figure}[htbp]
  \centering
  \includegraphics[width=\columnwidth]{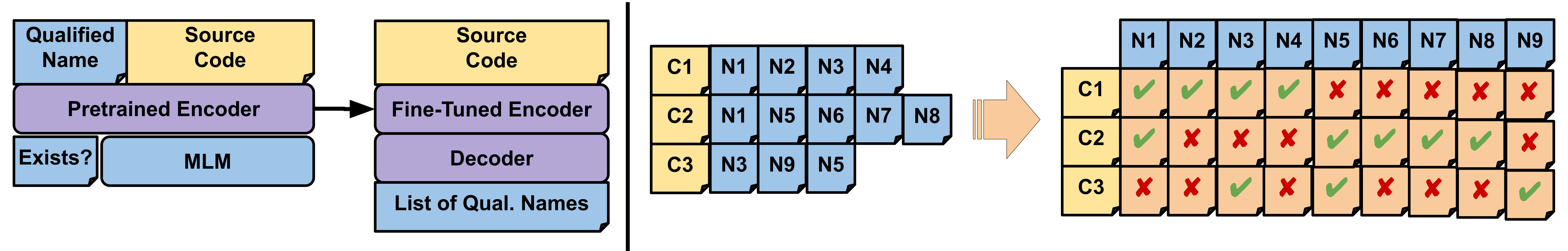}
  \caption{On the left, we train an encoder to predict whether a qualified name refers to a declaration in a given code. Then this encoder is attached to a decoder to form a full transformer which we train to generate the full list of declarations. On the right, we illustrate how a batch of inputs (code and list of qualified names) can easily be used to form contrastive pairs to train the encoder.  (See color legend in figure~\ref{fig:models})}
  \label{fig:scaml}
  \vspace{-.5\baselineskip}
\end{figure}

The first step of the pipeline for this task looks at fine-tuning a pretrained encoder to predict whether a qualified-name has a matching declaration in a code. There is a large number of paired inputs for this task. So we can easily create negative pairs to train the classification head of any classifier contrastively. The second step uses this fine-tuned encoder and trains a decoder to produce the full list of qualified names from C++ code.

\section{Related Work}
With the boom in machine learning applied to various research domains, many research activities survey and categorize the ML models, tasks, and the applied techniques.
Allamanis \textit{et al.} discuss how the similarities and differences between natural languages and program languages drive the design of probabilistic models.
A survey was provided to review how researchers adapted these probabilistic models to an application areas and discusses 
crosscutting and application-specific challenges and opportunities~\cite{allamanis2018survey}.
Maas introduced a taxonomy of ML for systems that aims to provide guidance, based on the proposed decision diagram, if machine learning should be applied to a particular system problem~\cite{9153088}.
A decision diagram is designed to provide recommendations to  practitioners and researchers to choose the most suitable machine learning strategies. 
Ashouri \textit{et al.} survey and classify recent research activities in compiler optimization, for optimization selection and phase ordering, with machine learning approaches~\cite{10.1145/3197978}. 
Kalyan \textit{et al.} summarize core concepts of transformer-based pretrained language models (T-PTLMs) and present taxonomy of T-PTLMs with brief overview of various benchmarks~\cite{https://doi.org/10.48550/arxiv.2108.05542}.
Sarker presents a deep learning taxonomy to cover techniques from major categories including supervised or discriminative learning, unsupervised or generative learning and hybrid learning~\cite{sarker2021deep}.
Chami \textit{et al.} present a taxonomy in graph representation learning (GRL)~\cite{https://doi.org/10.48550/arxiv.2005.03675} to include GRL from  network embedding, graph regularization and graph neural networks.  




Different from the cited research works, this paper surveys and creates a set of taxonomies for multiple components in machine learning to represent the PLP tasks, ML model architectures and the tokenization tools associated with the ML models.  In addition, this work demonstrates how to reference the taxonomies to assemble specialized PLP pipelines for new downstream machine learning tasks.

\section{Conclusion}
In this paper,  we have selected a set of representative papers in the domain of programming language processing (PLP) using machine learning. We have identified and categorized common PLP tasks and the associated reusable components so newcomers can easily 
find the right models and datasets for a given task. Using two example tasks, we have shown that the discovered components can be easily reused to construct customized machine learning pipelines to solve the given tasks.    

We started to implement the pipelines we have described above to learn more about DataRaceBench and construct challenging PLP problems. We will also encode the information into a formal knowledge representation such as ontology to enable automated pipeline adaptation using workflow synthesis techniques. 

\bibliographystyle{splncs04}
\bibliography{reference.bib,knowledge.bib,survey.bib}

\end{document}